\pdfoutput=1

\documentclass[11pt]{article}

\usepackage[preprint]{acl}

\usepackage{times}
\usepackage{latexsym}
\usepackage{booktabs}
\usepackage{multirow}
\usepackage{enumitem}

\usepackage[T1]{fontenc}

\usepackage[utf8]{inputenc}

\usepackage{microtype}

\usepackage{inconsolata}

\usepackage{graphicx}
\usepackage{amsmath}

\title{
\begin{tabular}{@{}l@{\hspace{0.2cm}} c@{}}
\multirow{2}{*}{\includegraphics[width=1.0cm]{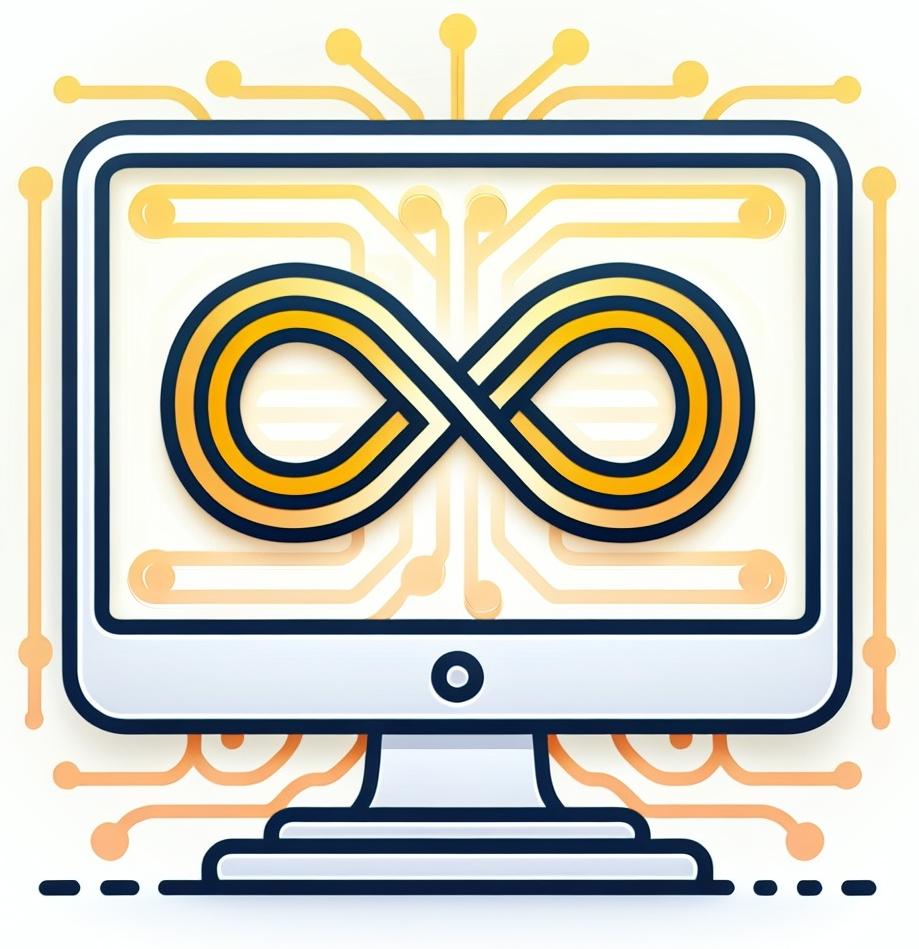}} & InfiGUIAgent: A Multimodal Generalist GUI Agent \\
 & with Native Reasoning and Reflection
\end{tabular}
}

\author{
 \textbf{Yuhang Liu\textsuperscript{1}},
 \textbf{Pengxiang Li\textsuperscript{2}},
 \textbf{Zishu Wei\textsuperscript{1}},
 \textbf{Congkai Xie\textsuperscript{3}},
 \textbf{Xueyu Hu\textsuperscript{1}},
 \textbf{Xinchen Xu\textsuperscript{1}},
\\
 \textbf{Shengyu Zhang\textsuperscript{1}},
 \textbf{Xiaotian Han \textsuperscript{4}},
 \textbf{Hongxia Yang\textsuperscript{5}},
 \textbf{Fei Wu\textsuperscript{1}}
\\
\\
 \textsuperscript{1}Zhejiang University,
 \textsuperscript{2}Dalian University of Technology,
 \textsuperscript{3}Reallm Labs,
\\
 \textsuperscript{4}ByteDance Inc,
 \textsuperscript{5}The Hong Kong Polytechnic University
\\
\small{
 \texttt{sy\_zhang@zju.edu.cn, xiaotian.han@bytedance.com, hongxia.yang@polyu.edu.hk}
 }
}

\begin{document}
\maketitle

\begin{abstract}
Graphical User Interface (GUI) Agents, powered by multimodal large language models (MLLMs), have shown great potential for task automation on computing devices such as computers and mobile phones. However, existing agents face challenges in multi-step reasoning and reliance on textual annotations, limiting their effectiveness. We introduce \textit{InfiGUIAgent}, an MLLM-based GUI Agent trained with a two-stage supervised fine-tuning pipeline. Stage 1 enhances fundamental skills such as GUI understanding and grounding, while Stage 2 integrates hierarchical reasoning and expectation-reflection reasoning skills using synthesized data to enable native reasoning abilities of the agents. \textit{InfiGUIAgent} achieves competitive performance on several GUI benchmarks, highlighting the impact of native reasoning skills in enhancing GUI interaction for automation tasks. Resources are
available at \url{https://github.com/Reallm-Labs/InfiGUIAgent}.
\end{abstract}
\section{Introduction}
Graphical User Interface (GUI) Agents have emerged as powerful tools for automating tasks on computing devices, including mobile phones and computers. These agents can understand and interact with GUIs to execute complex operations, significantly enhancing user productivity and expanding the scope of automated task completion \citep{202412.2294,hong2024cogagent,zhang2023you,qi2024webrl,xie2024osworld,vu2024gptvoicetasker,yu2024exact,wen2023autodroid}. 

Recent developments in multimodal large language models (MLLMs) \citep{Bai2023QwenVLAF,li2024ariaopenmultimodalnative,team2024gemini,dai2022modelmultiplemodalitiessparsely} have significantly advanced the potential of GUI Agents. MLLMs possess powerful visual understanding capabilities and can reason based on visual information, making them a promising foundation for building sophisticated GUI Agents. These models can interpret complex interface elements and adapt to a wide range of tasks, leading to more efficient and robust automation \citep{hong2024cogagent, jiang2023iluvui, you2025ferret, nong2024mobileflow, vu2024gptvoicetasker}.

However, current MLLM-based GUI Agents face several critical challenges. A key limitation lies in their reasoning capabilities \citep{zhang2023you, qi2024webrl, yu2024exact}. While many existing GUI Agents can perform basic single-step reasoning, they struggle to effectively leverage information from previous steps. This lack of reflection on past experiences can lead to repetitive errors during task execution. 

Another significant challenge lies in the reliance on the additional information of the GUIs. Many prior GUI Agent implementations rely on accessibility trees or Set-of-Marks \citep{yang2023setofmarkpromptingunleashesextraordinary}, to represent or augment the GUI's visual information. However, GUIs are inherently visual, and representing them primarily through text can lead to information loss or redundancy. Augmenting visual input with textual descriptions can also increase computational overhead. Furthermore, the availability and consistency of these textual representations vary across platforms, hindering practical deployment.

To address these limitations, we propose \textit{Infi\-GUIAgent}, which is a MLLM-based GUI Agent trained through a two-stage supervised fine-tuning (SFT) methods with robust fundamental capabilities and native reasoning abilities. In stage 1, we collect data covering multiple tasks, such as vision-language understanding, GUI-specific QA, and tool use to improve fundamental capabilities such as GUI understanding and instruction grounding of the agents. In stage 2, we recognized two essential reasoning skills for GUI Agents: (1) Hierarchical reasoning, and (2) Expectation-reflection reasoning, and integrate these skills into the SFT data synthesized by MLLMs based on existing trajectories. Our main contributions are threefold:
\begin{itemize}
    \item We propose a two-stage supervised fine-tuning pipeline to comprehensively improve both the fundamental abilities and advanced reasoning abilities of GUI Agents.
    \item We synthesize SFT data with two advanced reasoning skills: hierarchical reasoning and expectation-reflection reasoning, enabling the agents to natively perform complex reasoning. 
    \item We build \textit{Infi\-GUIAgent} by supervised fine-tuning a model using our SFT data and conduct experiments on several GUI benchmarks, demonstrating that our model achieves competitive performance.
\end{itemize}
\section{Related Works}
\subsection{Multimodal LLMs}
Large Language Models (LLMs) \citep{floridi2020gpt, Touvron2023LLaMAOA, Bai2023QwenTR,xiao2021lawformer} have significantly enhanced the capabilities of AI systems in tackling a wide range of tasks \citep{hu2024infiagentdabench, li2024inficodereval}, thanks to their exceptional ability to process complex semantic and contextual information. The remarkable power of LLMs has also inspired exploration into their potential for processing multimodal data, such as images. Typically, the architecture of Multimodal Large Language Models (MLLMs) consists of three main components: a pre-trained large language model, a trained modality encoder, and a modality interface that connects the LLM with the encoded modality features. Various vision encoders, such as ViT \citep{Dosovitskiy2020AnII}, CLIP \citep{radford2021learning}, and ConvNeXt \citep{liu2022convnet}, extract visual features, which are integrated using techniques like adapter networks \citep{Liu2023VisualIT}, cross-attention layers \citep{Alayrac2022FlamingoAV}, and visual expert modules \citep{wang2023cogvlm}. These methods have facilitated the development of high-performing MLLMs, such as Qwen-VL \citep{Bai2023QwenVLAF}, GPT-4 Vision \citep{openai2023gpt4v}, BLIP-2 \citep{li2023blip} and InfiMM \citep{liu2024infimm}, thus opening new avenues for LLMs in processing GUI tasks.

\subsection{MLLM-based GUI Agents}
Agents are AI systems that perceive their environments, make decisions, and take actions to complete specific tasks. LLMs reaching human-level intelligence have greatly enhanced the ability to build agents. For GUI tasks, LLMs that read HTML code to perceive GUIs are developed \citep{wen2023autodroid}. However, various works have shown that learning to interact with the visual form of the GUIs can show superior performance \citep{202412.2294}. Therefore, MLLM-based GUI Agents are developed. ILuvUI \citep{jiang2023iluvui} fine-tuned LLaVA to enhance general GUI understanding, while AppAgent \citep{zhang2023appagent} explored app usage through autonomous interactions. CogAgent \citep{hong2024cogagent} integrated high-resolution vision encoders, and Ferret-UI-anyres \citep{you2025ferret} employed an any-resolution approach. Building upon these works, our study focuses on developing a more lightweight agent with a simplified architecture for GUI tasks, aiming to improve ease of deployment.

\begin{figure*}[ht]
\centering
\includegraphics[width=\textwidth]{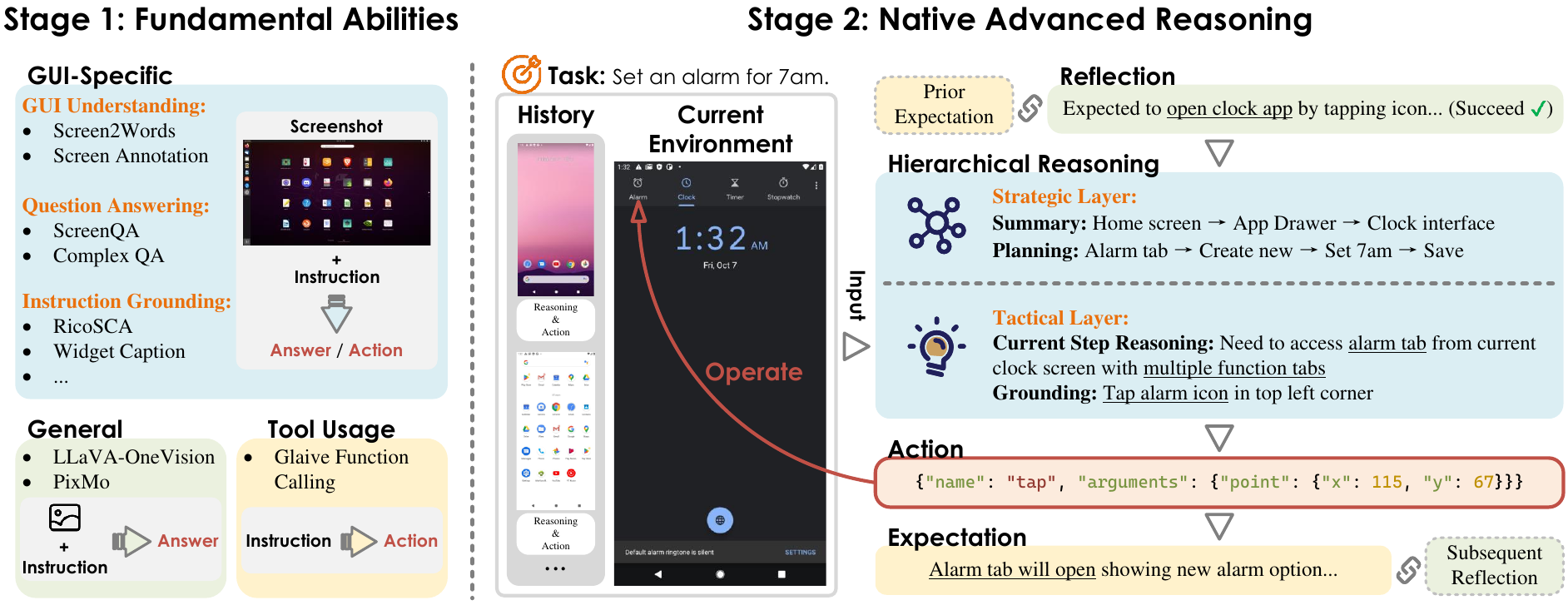}
\caption{
\textit{InfiGUIAgent} is trained in two stages. \textbf{Stage 1} cultivates fundamental abilities using diverse datasets covering GUI understanding (element recognition and layout comprehension), question answering, instruction grounding, general knowledge, and tool usage. \textbf{Stage 2} introduces native advanced reasoning, employed during both training and inference. This stage follows a cyclical process at each step, consisting of \texttt{Reflection}, \texttt{Hierarchical Reasoning} (strategic and tactical layers), \texttt{Action}, and \texttt{Expectation}. Each step receives the overall task, the history of previous screenshots and reasoning, and the current environment as input. \texttt{Reflection} assesses the previous action's outcome against its expectation, while \texttt{Expectation} predicts the outcome of the current action for subsequent reflection.
}
\label{fig:method}
\end{figure*}

\section{Method}
In this section, we introduce our two-stage supervised fine-tuning strategy for building \textit{InfiGUIAgent}, as shown in Figure~\ref{fig:method}. In stage 1, we focus on improving fundamental abilities such as understanding and grounding, particularly considering the complexity of GUIs. In stage 2, we move on to improve the native reasoning abilities of agents for handling complicated GUI tasks. 
 
\subsection{Stage 1: Training for Fundamental Abilities}
\begin{table*}
\centering
\small
\caption{Training datasets used in stage 1 of supervised fine-tuning.}
\label{tab:data_stage_one}
\begin{tabular}{lllr}
\toprule
\textbf{Dataset} & \textbf{Platform} & \textbf{Category} & \textbf{\# of Samples} \\
\midrule
\multicolumn{4}{l}{\textit{GUI-related Datasets}} \\
GUIEnv \citep{chen2024guicourse} & Webpage & Grounding & 150,000 \\
RICO Semantic Annotation \citep{sunkara2022towards} & Mobile & Grounding & 150,000 \\
SeeClick-Web \citep{cheng2024seeclick} & Webpage & Grounding & 100,000 \\
RICO SCA \citep{li2020ricosca} & Mobile & Grounding & 100,000 \\
Widget Caption \citep{li2020widget} & Mobile & Grounding & 70,000 \\
GUIChat \citep{chen2024guicourse} & Webpage & QA & 40,000 \\
ScreenQA \citep{hsiao2022screenqa} & Mobile & QA & 17,000 \\
UIBert Reference Expression \citep{bai2021uibert} & Mobile \& Mobile & Grounding & 16,000 \\
Screen2Words \citep{wang2021screen2words} & Mobile & Understanding & 12,000 \\
Complex QA \citep{yin2023lumos} & Mobile & QA & 11,000 \\
Screen Annotation \citep{baechler2024screenai} & Mobile & Understanding & 5,400 \\
OmniAct-Single Click \citep{kapoor2024omniact} & Webpage \& Desktop & Grounding & 4,800 \\
\midrule
\multicolumn{4}{l}{\textit{Non-GUI Datasets}} \\
LLaVA-OneVision \citep{llavaonevision} & - & General & 250,000 \\
PixMo \citep{allenai2024molmopixmo} & - & General & 68,800 \\
Glaive-function-calling \citep{glaive2024function} & - & Tool Usage & 5,000 \\
\bottomrule
\end{tabular}
\end{table*}

Considering the complexity of GUIs, which involve diverse data formats such as HTML code, high-resolution interfaces cluttered with small icons and text, general MLLMs lack fundamental abilities in both understanding GUI and grounding the actions. To address this, we first collected a range of existing visual-language and GUI datasets for supervised fine-tuning in stage 1. We gathered data covering several GUI tasks from multiple sources to ensure a comprehensive capabilities improvement (see Table~\ref{tab:data_stage_one}). The datasets can be categorized into five parts:

\begin{itemize}[leftmargin=*]
\setlength\itemsep{0em}
\setlength\parsep{0em}
\setlength\topsep{0em}
\setlength\parskip{0em}
\setlength\partopsep{0em}
    \item \textbf{GUI Understanding.} Datasets focusing on GUI element recognition, layout comprehension, and semantic interpretation, including Screen2Words \citep{wang2021screen2words} and Screen Annotation \citep{baechler2024screenai}.
    \item \textbf{Grounding.} Datasets capture various user interaction sequences and operation patterns, including GUIEnv \citep{chen2024guicourse}, RICO Semantic Annotation \citep{sunkara2022towards}, SeeClick-Web \citep{cheng2024seeclick}, RICO SCA \citep{li2020ricosca}, Widget Caption \citep{li2020widget}, UIBert Reference Expression \citep{bai2021uibert} and OmniAct-Single Click \citep{kapoor2024omniact}.
    \item \textbf{Question Answering.} Datasets contain GUI-specific QA tasks, including GUIChat \citep{chen2024guicourse}, ScreenQA \citep{hsiao2022screenqa} and Complex QA \citep{yin2023lumos}.
    \item \textbf{General Knowledge.} Multimodal datasets maintain model's general capabilities, including LLaVA-OneVision \citep{llavaonevision} and PixMo \citep{allenai2024molmopixmo}.
    \item \textbf{Tool Usage.} Datasets cover general tool using, including Glaive-function-calling \citep{glaive2024function}.
\end{itemize}

Due to the diversity of our data sources, we implemented comprehensive format standardization across all datasets. Additionally, we adopted the Reference-Augmented Annotation format (see Section~\ref{sec:raa}) to enhance the model's ability to ground visual elements with textual descriptions, enabling precise spatial referencing while maintaining natural language flow.

\subsubsection{Data Preprocessing and Standardization}
Given the diversity of our data sources, we implemented comprehensive preprocessing steps to standardize the data format across all datasets. We normalized the coordinate system by following \cite{wang2024qwen2}, mapping all spatial coordinates to a relative scale of [0, 1000].
This standardization facilitates consistent representation of both point and box annotations in JSON format, with points expressed as \(\{\text{"x"}: x, \text{"y"}: y\}\) and bounding boxes as \(\{\text{"x1"}: x_1, \text{"y1"}: y_1, \text{"x2"}: x_2, \text{"y2"}: y_2\}\). In this coordinate system, the origin \(\{\text{"x"}: 0, \text{"y"}: 0\}\) is located at the screen's top-left corner, with the x-axis extending rightward and the y-axis downward. The bottom-right corner corresponds to coordinates \(\{\text{"x"}: 1000, \text{"y"}: 1000\}\). To enhance data quality, we implemented two additional preprocessing steps:
\begin{itemize}[leftmargin=*]
\setlength\itemsep{0em}
    \item \textbf{Instruction Enhancement.} For datasets with ambiguous instructions, we developed standardized instruction templates to establish clear correspondence between commands and their expected outcomes.
    \item \textbf{Response Refinement.} For entries with complex or inconsistent response formats, we utilized Qwen2-VL-72B \citep{Bai2023QwenVLAF} to reformulate responses while preserving their semantic content. Each reformulation underwent validation to ensure accuracy and consistency.
\end{itemize}

\subsubsection{Reference-Augmented Annotation} \label{sec:raa}
To better leverage the spatial information available in our collected datasets and enhance the model's visual-language understanding of GUIs, we implemented a reference-augmented annotation format. This format enables bidirectional referencing between GUI elements and textual responses. Specifically, we adopted the following structured notation:

\begin{figure}[h]
\includegraphics[width=0.35\textwidth]{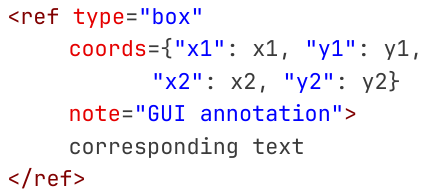}
\end{figure}

The format consists of several key components: the reference type (either "box" for rectangular regions or "point" for specific locations), coordinate specifications (x1, y1, x2, y2 for boxes or x, y for points), optional annotative notes, and the corresponding textual content. To generate training data in this format, we prompted Qwen2-VL-72B \citep{Bai2023QwenVLAF} to seamlessly integrate GUI spatial information with original responses, maintaining natural language flow while preserving precise spatial references.

\subsection{Stage 2: Training for Native Reasoning}
\begin{table*}
\centering
\small
\caption{UI action reasoning datasets used in the training process}
\label{tab:data_stage_two}
\begin{tabular}{llr}
\toprule
\textbf{Dataset} & \textbf{Platform} & \textbf{\# of Samples} \\
\midrule
GUIAct \citep{chen2024guicourse} & Webpage \& Mobile & 10,000\\
AMEX \citep{amex} & Mobile & 3,000\\
Android in the Zoo \citep{zhang2024android} & Mobile & 2,000\\
Composition: Stage 1-aligned & - & 30,000\\
\bottomrule
\end{tabular}
 
\end{table*}

Building upon the foundational capabilities such as understanding and grounding, GUI Agents must also master advanced reasoning skills to effectively handle complex tasks. We identify two crucial reasoning skills : (1) Hierarchical reasoning, which enables planning and task decomposition, helping agents structure complex tasks into manageable subtasks and execute them efficiently \citep{huang2023reasoninglargelanguagemodels,zhang2024llmmastermindsurveystrategic,huang2024understanding}, and (2) Expectation-reflection reasoning, which fosters adaptive self-correction and reflection \citep{shinn2023reflexionlanguageagentsverbal,yao2023reactsynergizingreasoningacting,hu2024leveragingprintdebuggingimprove}, enabling agents to learn from past actions and improve decision-making consistency. These reasoning skills are integrated into the training datasets of agents, so that they can reason with these skills natively without any extra prompting. To achieve this, we generate SFT data incorporating these reasoning skills based on existing trajectory data (see Table~\ref{tab:data_stage_two}) and continue fine-tuning the model from stage 1.

\subsubsection{Hierarchical Reasoning}
Effective execution of GUI tasks demands both overarching strategic planning and meticulous tactical execution. To achieve this, we synthesize trajectory data with a hierarchical reasoning with two distinct layers:
\begin{itemize}[leftmargin=*]
\setlength\itemsep{0em}
\setlength\parsep{0em}
\setlength\topsep{0em}
\setlength\parskip{0em}
\setlength\partopsep{0em}
    \item \textbf{Strategic Layer.} Strategic layer is responsible for high-level task decomposition and sub-goal planning. This layer analyzes the overall task objective and determines the sequence of subtasks needed for completion.
    \item \textbf{Tactical Layer.} Tactical layer handles the selection and grounding of concrete actions. Based on the strategic layer's planning, agent select appropriate GUI operations and adjusts their parameters to match the target.
\end{itemize}

\subsubsection{Expectation-Reflection Reasoning}
To enhance action consistency and foster autonomous self-correction, we incorporate Expectation-reflection reasoning into the training datasets. This iterative process enhances the agent's ability to adapt and learn from its actions through a structured reflection cycle:
\begin{itemize}[leftmargin=*]
\setlength\itemsep{0em}
\setlength\parsep{0em}
\setlength\topsep{0em}
\setlength\parskip{0em}
\setlength\partopsep{0em}
    \item \textbf{Reasoning.} After reflection (except the first step), the agents conduct hierarchical reasoning.
    \item \textbf{Action.} After the reasoning, the agent takes the action.
    \item \textbf{Expectation.} Following each action, the agent generates expected outcomes which are used to be verified at the next step.
    \item \textbf{Reflection.} The agent evaluates whether its actions achieved the expected results and generating a textual summary of the reflection.
\end{itemize}

\begin{table}[t]
\centering
\resizebox{\columnwidth}{!}{%
\begin{tabular}{ll}
\toprule
\textbf{Category} & \textbf{Operations} \\
\midrule
\textbf{Single-point operations} & \texttt{tap}, \texttt{click}, \texttt{hover}, \texttt{select} \\
\textbf{Two-point operations} & \texttt{swipe}, \texttt{select\_text} \\
\textbf{Directional operations} & \texttt{scroll} \\
\textbf{Text input} & \texttt{input}, \texttt{point\_input} \\
\textbf{Parameterless operations} & \texttt{remember}, \texttt{enter}, \texttt{home}, \texttt{back} \\
\textbf{State settings} & \texttt{set\_task\_status} \\
\bottomrule
\end{tabular}%
}
\caption{Categorization of actions in the action space.}
\label{tab:action_space}
\end{table}

\subsubsection{Agent-Environment Interface}
We formulate the GUI interaction as a process where an agent interacts with a mobile environment. Let $s_t \in \mathcal{S}$ denote the environment state at step $t$, where $\mathcal{S}$ represents the state space. The agent can observe the state through a screenshot observation $o_t$ and performs actions $a_t \in \mathcal{A}$, where $\mathcal{A}$ is the action space. The environment transitions from $s_t$ to $s_{t+1}$ following $s_{t+1} \sim P(\cdot|s_t, a_t)$, where $P$ represents the transition probability function.

The agent receives a task goal $g$ and maintains access to a history window of size $n$. At each step $t$, the agent's input consists of:
\begin{itemize}[leftmargin=*]
\setlength\itemsep{0em}
    \item Goal $g$
    \item Current observation $o_t$
    \item Historical context $H_t = \{(o_i, r_i, a_i)\}_{i=t-n}^{t-1}$, where $r_i$ represents the reasoning process
\end{itemize}

Based on these inputs, the agent generates a reasoning process $r_t$ and predicts an action $a_t$. The interaction follows a standard protocol using function calls and responses:

\begin{figure}[h]
\includegraphics[width=0.40\textwidth]{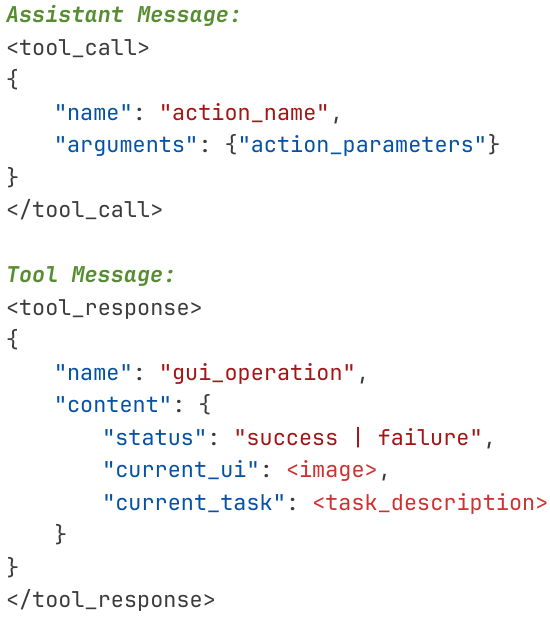}
\end{figure}

\subsubsection{Modular Action Space}

Given the diverse action spaces across collected datasets, we categorized and standardized the actions by unifying their names and parameters, merging similar operations where appropriate. The resulting action space $\mathcal{A}$ consists of independent, composable operations that can be flexibly combined based on task requirements, as shown in Table~\ref{tab:action_space}. 
This modular design allows for dynamic action space configuration while maintaining a consistent interface across different platforms and scenarios.

\subsubsection{Reasoning Process Construction}
To construct high-quality reasoning data to stimulate the model's native reasoning capabilities, we leverage more capable MLLMs (e.g. Qwen2-VL-72B) to generate structured reasoning processes based on existing interaction trajectories. The construction process involves several key components:

\begin{itemize}[leftmargin=*]
\setlength\itemsep{0em}
    \item \textbf{Screenshot Description.} For each observation $o_t$ in the trajectory, we generate a detailed description $d_t$.
    This step addresses the limitation that some MLLM models do not support interleaved image-text input formats well. To establish clear correspondence between observations (screenshots) and steps, we generate detailed descriptions to replace the screenshots, which helps facilitate the subsequent reasoning process construction.
    
    \item \textbf{Reflection.} Given the previous expectation $e_{t-1}$ and current observation $o_t$, we generate a reflection $f_t$ that evaluates the outcome of the previous action.
    
    \item \textbf{Strategic Layer.} The strategic reasoning consists of two parts: First, a summary is generated based on the n-step history $H_t = \{(o_i, r_i, a_i)\}_{i=t-n}^{t-1}$ and current observation $o_t$. Then, the planning component is generated with access to the actual action $a_t$ to ensure alignment with the trajectory.
    
    \item \textbf{Tactical Layer.} This layer's reasoning is constructed using the generated reflection $f_t$ and strategic layer output. The actual action $a_t$ from the trajectory is incorporated to ensure the tactical reasoning leads to appropriate action selection.
    
    \item \textbf{Expectation.} For each state-action pair $(s_t, a_t)$, we generate an expectation $e_t$ based on current observation $o_t$, reasoning process $r_t$, and action $a_t$. 
    Notably, we deliberately avoid using the next state $s_{t+1}$ in this generation process. Although using $s_{t+1}$ could improve the agent's accuracy in modeling state transitions, while using $s_{t+1}$ could lead to perfect expectations, such an approach might impair the agent's ability to handle expectation mismatches during deployment.
\end{itemize}

While we avoid using $s_{t+1}$ in expectation generation to maintain robustness, we also explore the possibility of improving state transition modeling through a parallel next-state prediction task. Using the trajectory data, we construct additional training examples where the agent learns to predict the next state description $d_{t+1}$ given the current observation $o_t$ and action $a_t$. 
This auxiliary task helps the agent learn state transition dynamics, while keeping the expectation generation process independent of future states.

\section{Experiments}

\subsection{Experimental Setting}
\begin{table*}[t]
\centering
\small
\begin{tabular}{l ccccccc}
\toprule
\multirow{2}{*}{\textbf{Model}} & \multicolumn{6}{c}{\textbf{Accuracy (\%)}} & \multirow{2}{*}{\textbf{Avg.}} \\
\cmidrule(lr){2-7}
& \multicolumn{2}{c}{Mobile} & \multicolumn{2}{c}{Desktop} & \multicolumn{2}{c}{Web} & \\
\cmidrule(lr){2-3} \cmidrule(lr){4-5} \cmidrule(lr){6-7}
& Text & Icon & Text & Icon & Text & Icon & \\
\midrule
\multicolumn{8}{l}{\textit{Proprietary Models}} \\
GPT-4o\footnote{gpt-4o-2024-08-06 version} \citep{gpt4o} & 30.5 & 23.2 & 20.6 & 19.4 & 11.1 & 7.8 & 18.8 \\
Gemini-1.5-pro\footnote{gemini-1.5-pro-002 version} \citep{team2024gemini} & 76.2 & 54.1 & 65.5 & 39.3 & 52.2 & 32.0 & 53.2 \\
\midrule
\multicolumn{8}{l}{\textit{Open-source Models}} \\
Qwen2-VL-2B \citep{wang2024qwen2} & 24.2 & 10.0 & 1.4 & 9.3 & 8.7 & 2.4 & 9.3 \\
Qwen2-VL-7B \citep{wang2024qwen2} & 61.3 & 39.3 & 52.0 & 45.0 & 33.0 & 21.8 & 42.9 \\
CogAgent \citep{hong2024cogagent} & 67.0 & 24.0 & 74.2 & 20.0 & 70.4 & 28.6 & 47.4 \\
SeeClick \citep{cheng2024seeclick} & 78.0 & 52.0 & 72.2 & 30.0 & 55.7 & 32.5 & 53.4 \\
UGround-7B \citep{gou2024navigating} & 82.8 & 60.3 & \underline{82.5} & 63.6 & \underline{80.4} & \textbf{70.4} & 73.3 \\
ShowUI-2B \citep{lin2024showui} & \textbf{92.3} & \textbf{75.5} & 76.3 & 61.1 & \textbf{81.7} & 63.6 & \underline{75.1} \\
\midrule
\multicolumn{8}{l}{\textit{Ours}} \\
\textbf{InfiGUIAgent-2B} & \underline{88.6} & \underline{74.7} & \textbf{85.6} & \textbf{65.0} & 79.1 & \underline{64.6} & \textbf{76.3} \\
\bottomrule
\end{tabular}
\caption{Performances on various platforms (Mobile, Desktop, Web) on Screenshot. All experiments were conducted using raw screenshot information. Results marked in \textbf{bold} represent the best performance, and those \underline{underlined} indicate the second-best performance.}
\label{table:comparison}
\end{table*}
\begin{table}[t]
\centering
\resizebox{\columnwidth}{!}{%
\begin{tabular}{l cccc}
\toprule
\multirow{2}{*}{\textbf{Model}} & \multicolumn{4}{c}{\textbf{Success Rate}} \\
\cmidrule(lr){2-5}
& Easy & Middle & Hard & Overall \\
\midrule
Qwen2-VL-2B \citep{wang2024qwen2} & 0.00 & 0.00 & 0.00 & 0.00 \\
Qwen2-VL-7B \citep{wang2024qwen2} & 0.00 & 0.00 & \textbf{0.05} & 0.05 \\
Qwen2-VL-72B \citep{wang2024qwen2} & 0.08 & 0.00 & \textbf{0.05} & 0.05 \\
LLaVa-OV-7B \citep{li2024llavaonevisioneasyvisualtask} & 0.00 & 0.00 & 0.00 & 0.00 \\
ShowUI-2B \citep{lin2024showui} & 0.18 & 0.00 & 0.00 & 0.07 \\
\midrule
\multicolumn{5}{l}{\textit{Ours}} \\
\textbf{InfiGUIAgent-2B} & \textbf{0.25} & 0.00 & 0.00 & \textbf{0.09} \\
\bottomrule
\end{tabular}%
}
\caption{Performances on AndroidWorld.}
\label{M3A}
\end{table}

\subsubsection{Implementation Details}
In stage 1, we sample 1M samples in total as illustrated in Table \ref{tab:data_stage_one}. In stage 2, we synthesized 45K samples based on trajectories from datasets shown in Table \ref{tab:data_stage_two}. We continual supervised fine-tune Qwen2-VL-2B \citep{bai2023qwen}. We leverage ZeRO0 \citep{rajbhandari2020zeromemoryoptimizationstraining} technology to enable full parameter fine-tuning of the model across 8 A800 80GB GPUs.
\subsubsection{Evaluation Benchmarks}
\paragraph{ScreenSpot.}
ScreenSpot \citep{cheng2024seeclick} is an evaluation benchmark for GUI grounding, consisting of over 1,200 instructions from iOS, Android, macOS, Windows, and Web environments, with annotated element types.

\paragraph{AndroidWorld.}
AndroidWorld \citep{rawles2024androidworld} is a fully functional Android environment that provides reward signals for 116 programmatic tasks across 20 real-world Android apps. We find that Android World uses Set-of-Marks (SoM)~\cite{yang2023set} to enhance the agent's grounding ability. However, when humans operate smartphones, their brains do not label elements on the screen. Over-reliance on SoM can lead to insufficient focus on pixel-level grounding ability. Therefore, in our experiments, agents respond to the raw image rather than the annotated image.

\subsection{Main Results}

\paragraph{ScreenSpot.}
Table~\ref{table:comparison} provides the results of different models across three platforms (Mobile, Desktop and Web) and two element types (Text and Icon) on ScreenSpot \citep{cheng2024seeclick}. InfiGUIAgent-2B achieves highest accuracy of 76.3\%, surpassing several strong baselines such as ShowUI \citep{lin2024showui} (75.1\%) and UGround-7B \citep{gou2024navigating} (73.3\%), which is even with larger parameters size.

\paragraph{AndroidWorld.}
Table~\ref{M3A} compares the success rates of \textit{InfiGUIAgent} with open-source models on AndroidWorld \citep{rawles2024androidworld}.  
InfiGUIAgent-2B achieves an overall success rate of 0.09, outperforming open-source models of similar size, such as ShowUI-2B \citep{lin2024showui} (0.07), and model with much more parameters such as LLaVa-OV-7B \citep{li2024llavaonevisioneasyvisualtask} (0.00) and Qwen2-VL-72B \citep{Bai2023QwenVLAF} (0.05).
\section{Conclusion}
In this work, we propose \textit{InfiGUIAgent}, a novel MLLM-based GUI Agents. By constructing comprehensive training datasets with two-stage supervised fine-tuning, we enhance the model's ability to understand, reason, and interact with GUIs. Our evaluation, conducted using raw screenshots without relying on additional GUI metadata, demonstrates the model's applicability to real-world scenarios. Experimental results show that our model performs well on GUI tasks and surpass several open-source baselines.

\bibliography{custom}

\appendix

\newpage
\section{Cases}\label{cases}

\subsection{Stage 1: Fundamental Abilities}
\begin{figure*}[ht]
    \centering
    \includegraphics[width=1.0\textwidth]{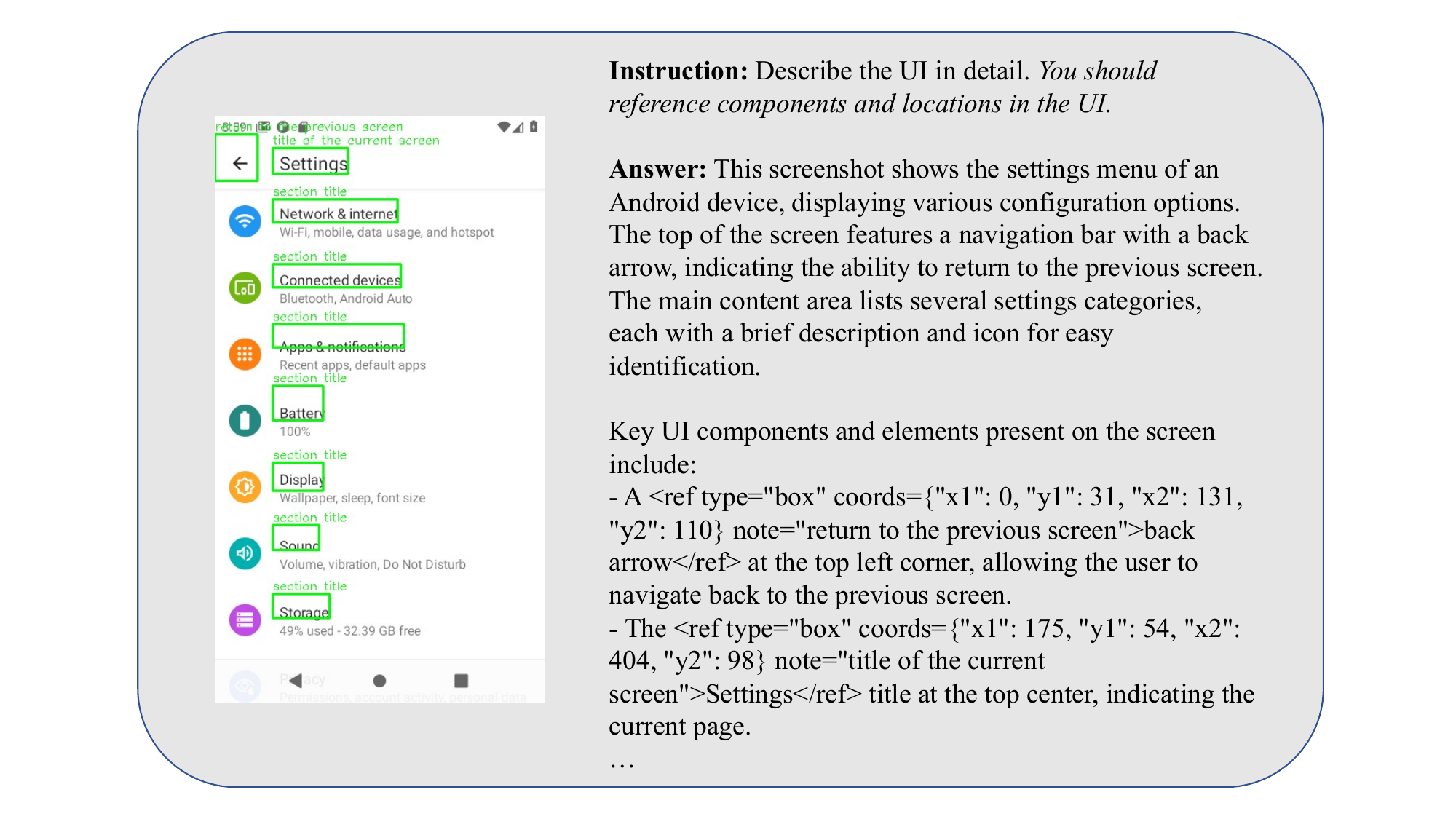} 
    \caption{Case of GUI Understanding. }
    \label{fig:case_understanding}
\end{figure*}

\begin{figure*}[ht]
    \centering
    \includegraphics[width=1.0\textwidth]{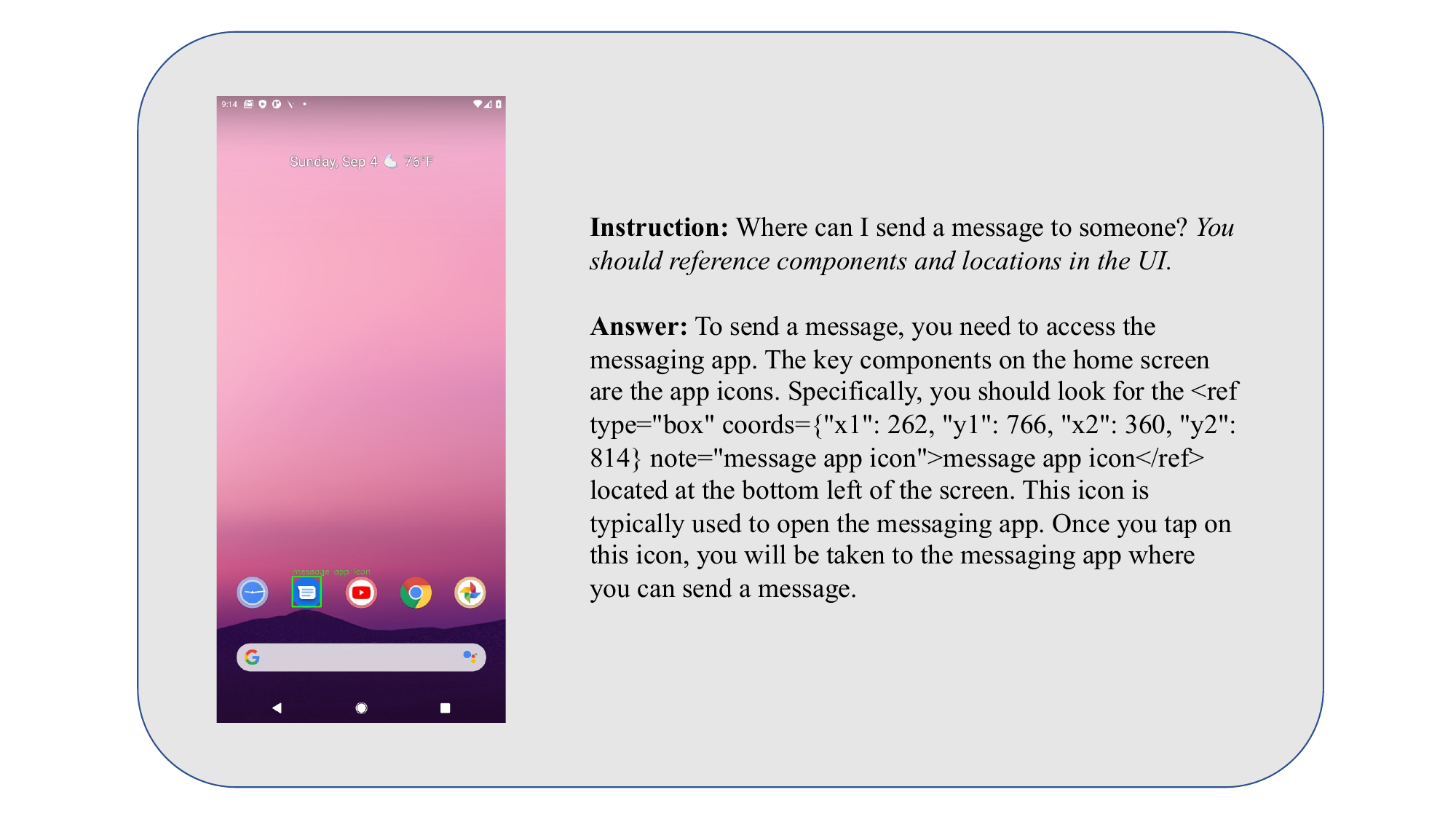} 
    \caption{Case of Grounding. }
    \label{fig:case_grounding}
\end{figure*}

\begin{figure*}[ht]
    \centering
    \includegraphics[width=1.0\textwidth]{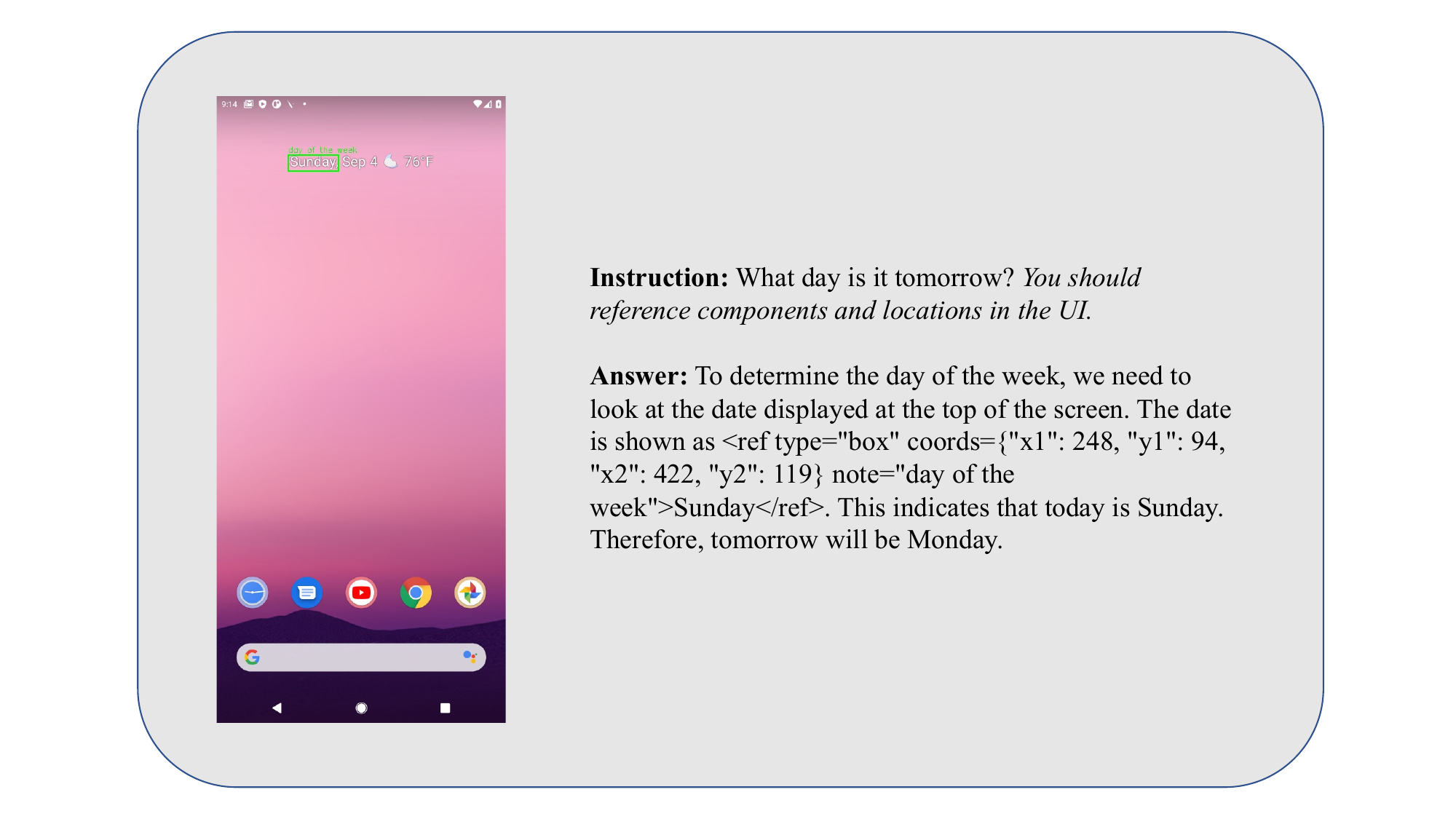} 
    \caption{Case of Question Answering. }
    \label{fig:case_qa}
\end{figure*}

We demonstrate the fundamental abilities trained in Stage 1 through three cases: GUI Understanding (Figure~\ref{fig:case_understanding}), Grounding (Figure~\ref{fig:case_grounding}), and Question Answering (Figure~\ref{fig:case_qa}).

\subsection{Stage 2: Native Reasoning}
\begin{figure*}[t]
    \centering
    \includegraphics[width=1.0\textwidth]{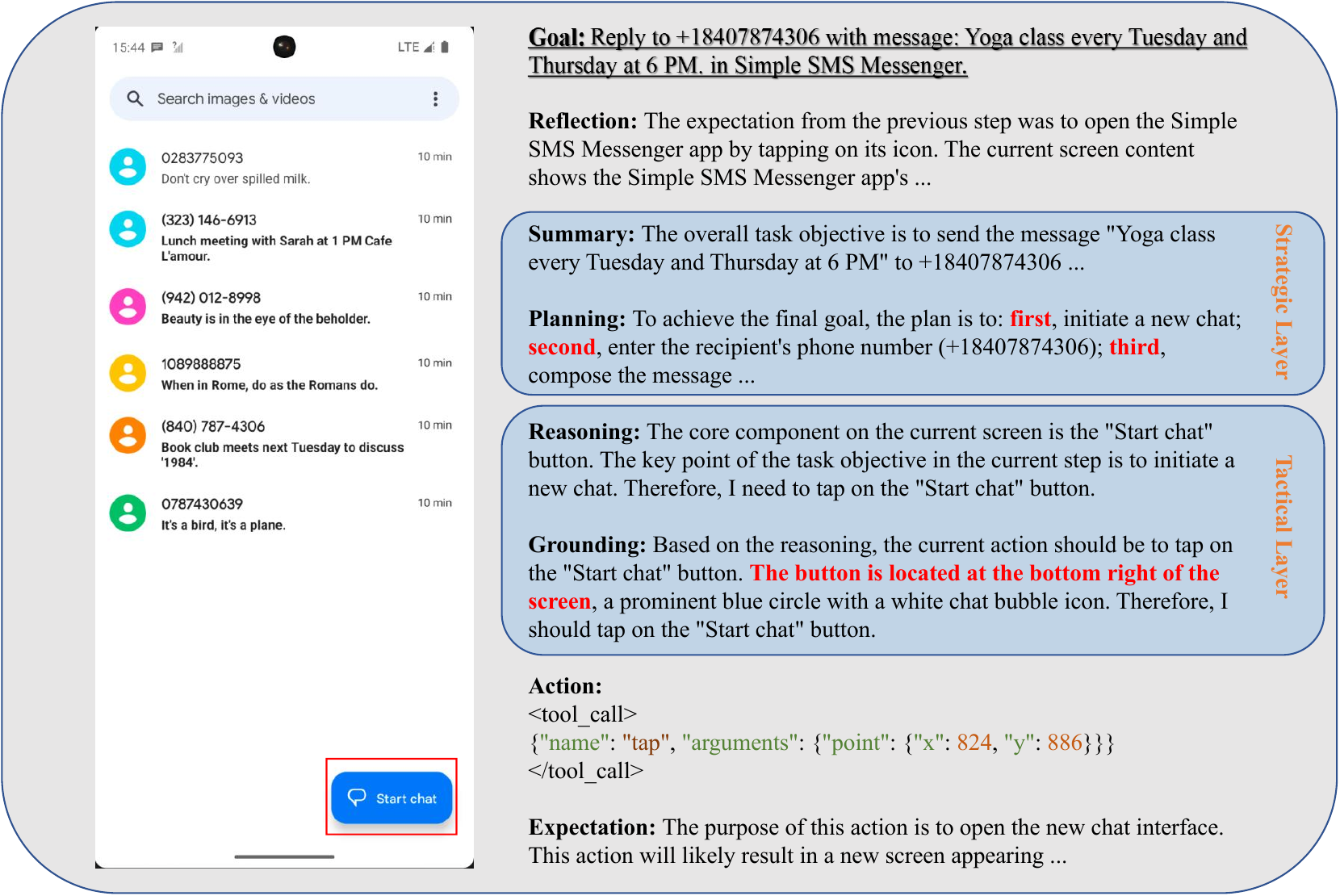} 
    \caption{Case of Native Advanced Reasoning. The agent's goal is to reply to a message}
    \label{fig:case_r3}
\end{figure*}
\begin{figure*}[t]
    \centering
    \includegraphics[width=1.0\textwidth]{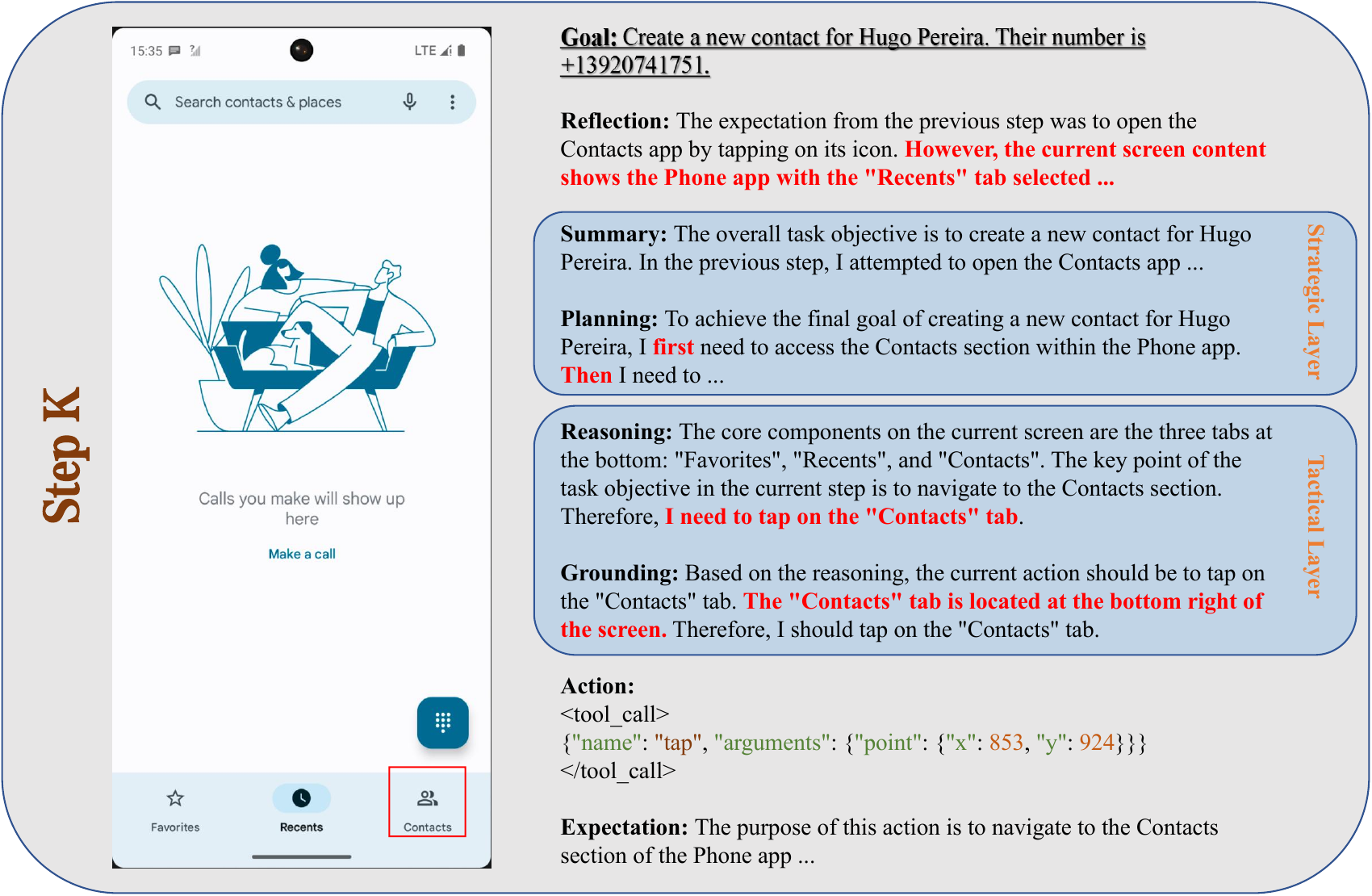} 
    \caption{Case of Native Advanced Reasoning. The agent's goal is to create a new contact.}
    \label{fig:case_r1}
\end{figure*}
\begin{figure*}[t]
    \centering
    \includegraphics[width=1.0\textwidth]{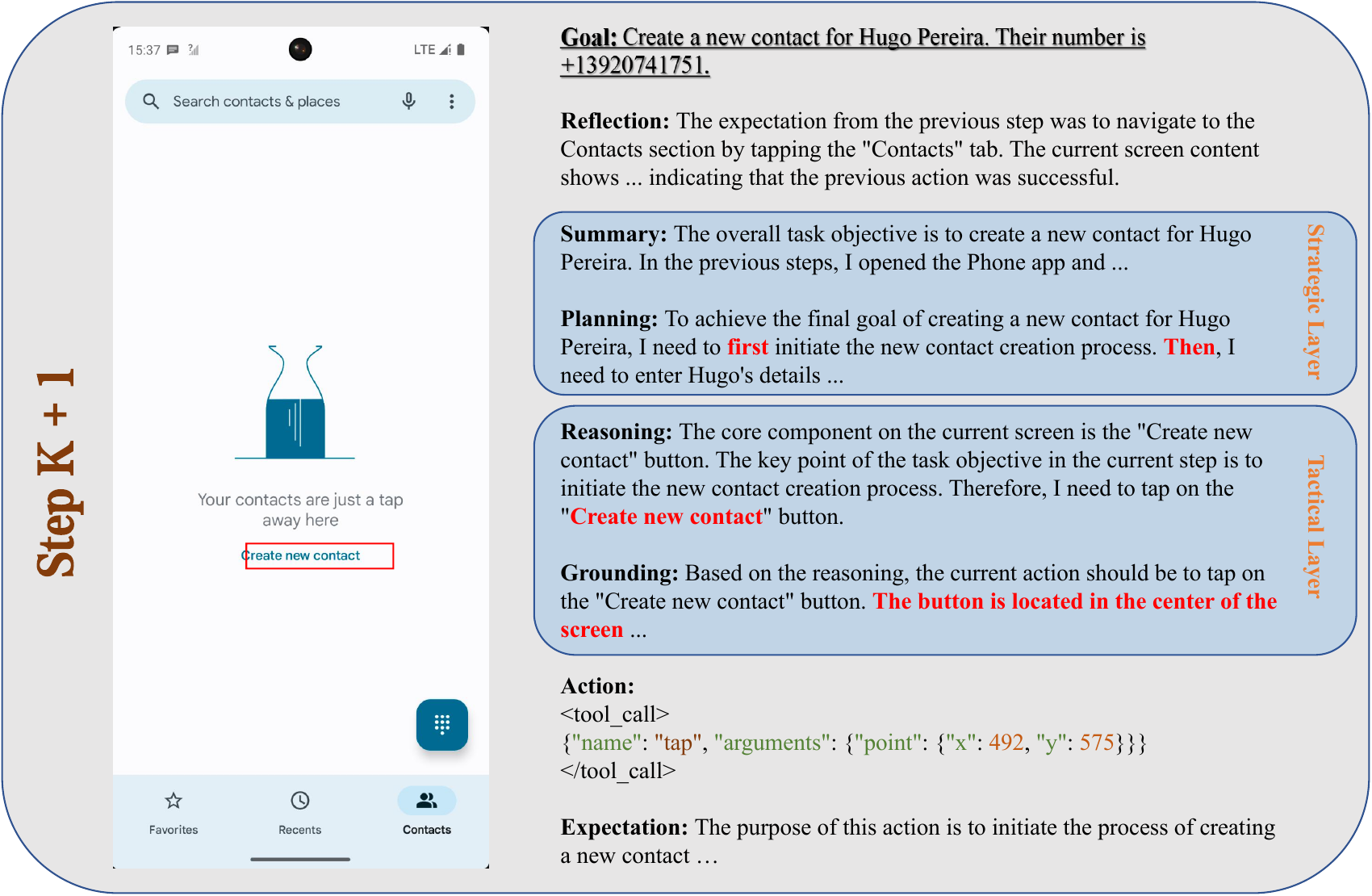} 
    \caption{Case of Native Advanced Reasoning. The agent's goal is to create a new contact.}
    \label{fig:case_r2}
\end{figure*}

We provide two representative cases to demonstrate the reasoning and interaction process of \textit{InfiGUIAgent}.

\paragraph{Reply to a Message} Figure~\ref{fig:case_r3} illustrates a step where the agent needs to reply to a specific message in a messaging application. The reasoning process involves identifying the "Start chat" button and grounding the action to initiate the reply process.

\paragraph{Creating a New Contact} Figure~\ref{fig:case_r1} and Figure~\ref{fig:case_r2} demonstrate sequential steps for creating a new contact. In the first step (Step K), the agent navigates to the "Contacts" section by reasoning and grounding the action to the corresponding tab. In the following step (Step K+1), the agent initiates the contact creation process by identifying and tapping the "Create new contact" button. These sequential steps highlight the agent's hierarchical reasoning and grounding abilities.

\end{document}